\documentclass{article} 
\usepackage{collas2022_conference,times}

\usepackage{amsmath,amsfonts,bm}

\def\eqref#1{equation~\ref{#1}}

\def\1{\bm{1}}

\DeclareMathAlphabet{\mathsfit}{\encodingdefault}{\sfdefault}{m}{sl}
\SetMathAlphabet{\mathsfit}{bold}{\encodingdefault}{\sfdefault}{bx}{n}

\usepackage{hyperref}
\hypersetup{
    colorlinks=true,
    linkcolor=red,
    filecolor=magenta,      
    urlcolor=blue,
    pdftitle={Overleaf Example},
    pdfpagemode=FullScreen,
    }

\usepackage{cite}
\usepackage{amsmath,amssymb,amsfonts}
\usepackage{algorithmic}
\usepackage{graphicx,subcaption}
\usepackage{textcomp}
\usepackage{xcolor}
\usepackage{hyperref}
\usepackage{todonotes}
\usepackage{rotating}
\usepackage[latin9]{inputenc}
\usepackage{tabularx}
\usepackage{soul}
\usepackage{booktabs}
\usepackage{multicol}
\usepackage{multirow}
\usepackage{amsmath}
\usepackage[bottom]{footmisc} 
\usepackage{wrapfig}

\usepackage{graphicx}

\usepackage{lipsum}
\usepackage{float}

\usepackage{xcolor,colortbl}
\definecolor{red}{rgb}{1.00,0.00,0.00}
\definecolor{blue}{rgb}{0.00,0.00,1.00}

\usepackage{url}
\def\BibTeX{{\rm B\kern-.05em{\sc i\kern-.025em b}\kern-.08em
    T\kern-.1667em\lower.7ex\hbox{E}\kern-.125emX}}

\begin{document}
\title{{Sim-To-Real Transfer of Visual Grounding for \\ Human-Aided Ambiguity Resolution}}
\author{Georgios Tziafas \\
Department of Artificial Intelligence\\
University of Groningen\\
The Netherlands \\
{\tt\small g.t.tziafas@rug.nl}
\And 
Hamidreza Kasaei \thanks{We thank NVIDIA Corporation for their generous donation of GPUs which was partially used in this research.} \\
Department of Artificial Intelligence\\
University of Groningen\\
The Netherlands \\
{\tt\small hamidreza.kasaei@rug.nl}
}
\collasfinalcopy
\maketitle

\begin{abstract}
Service robots should be able to interact naturally with non-expert human users, not only to help them in various tasks, but also to receive guidance in order to resolve ambiguities that might be present in the instruction. We consider the task of visual grounding, where the agent segments an object from a crowded scene given a natural language description. Modern holistic approaches to visual grounding usually ignore language structure and struggle to cover generic domains, therefore relying heavily on large datasets.
 Additionally, their transfer  performance in RGB-D datasets suffers due to high visual discrepancy between the benchmark and the target domains. Modular approaches marry learning with domain modeling and exploit the compositional nature of language to decouple visual representation from language parsing, but either rely on external parsers or are trained in an end-to-end fashion due to the lack of strong supervision. In this work, we seek to tackle these limitations by introducing a fully decoupled modular framework for compositional visual grounding of entities, attributes and spatial relations. 
 We exploit rich scene graph annotations generated in a synthetic domain and train each module independently in simulation.  Our approach is evaluated both in simulation and in two real RGB-D scene datasets. Experimental results show that the decoupled nature of our framework allows for easy integration with domain adaptation approaches for Sim-To-Real visual recognition, offering a data-efficient, robust, and interpretable solution to visual grounding in robotic applications.

\end{abstract}
\section{Introduction}
\label{intro}

\begin{wrapfigure}{r}{0.5\textwidth}
    \vspace{-4mm}
    \centering
    \includegraphics[width=\linewidth, trim = 4mm 0mm 14mm 2mm, clip=true]{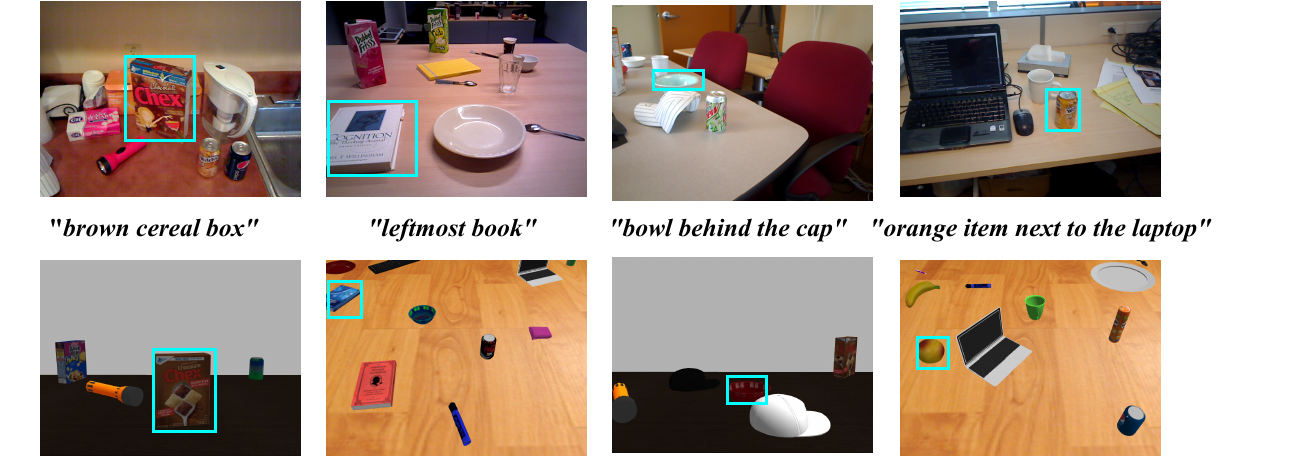}
    \caption{ Disambiguation of objects according to natural language query, including (\textit{from left to right}): reference to visual attribute (color), reference to location, reference to spatial relation with another object and combined. Examples are included from a real RGB-D scene \textit{(top-row)}, as well as from synthetically generated scenes \textit{(bottom-row)}.}
    \label{fig:intro}
\end{wrapfigure}
Natural Human-Robot Interaction (HRI) remains as one of the most valuable assets for service robots operating in human-centric environments. In such dynamic environments, the robot needs to be able to receive verbal input from non-expert users in order to coordinate its perception and manipulation functions (e.g. user says: \textit{``Give me the mug''}). The user should be also encouraged to assist the robot's perception in resolving ambiguities (e.g. mug appears twice in the scene), by querying for visual (e.g. color, i.e \textit{``the red mug''}) or spatial (\textit{``the mug next to the flashlight''}) concepts, as shown in Figure~\ref{fig:intro}. 
Most deployed robots miss this capability by treating their perception and HRI modules separately, often hard-coding maps from verbal instructions to specific object/scene setups. 
Such methods suffer from the limitation of not enabling \textit{natural} interaction with non-expert users (e.g. given object query does not appear in the hard-coded object set) and the need to be explicitly re-programmed for each new added object category or way of referring to an object. 
A bridging of the two modules is viable through end-to-end multi-modal learning models that are trained in image-text pairs.
\begin{figure*}[t]
    \centering
    \includegraphics[width=1\textwidth, trim = 5mm 10mm 17mm 0mm, clip=true]{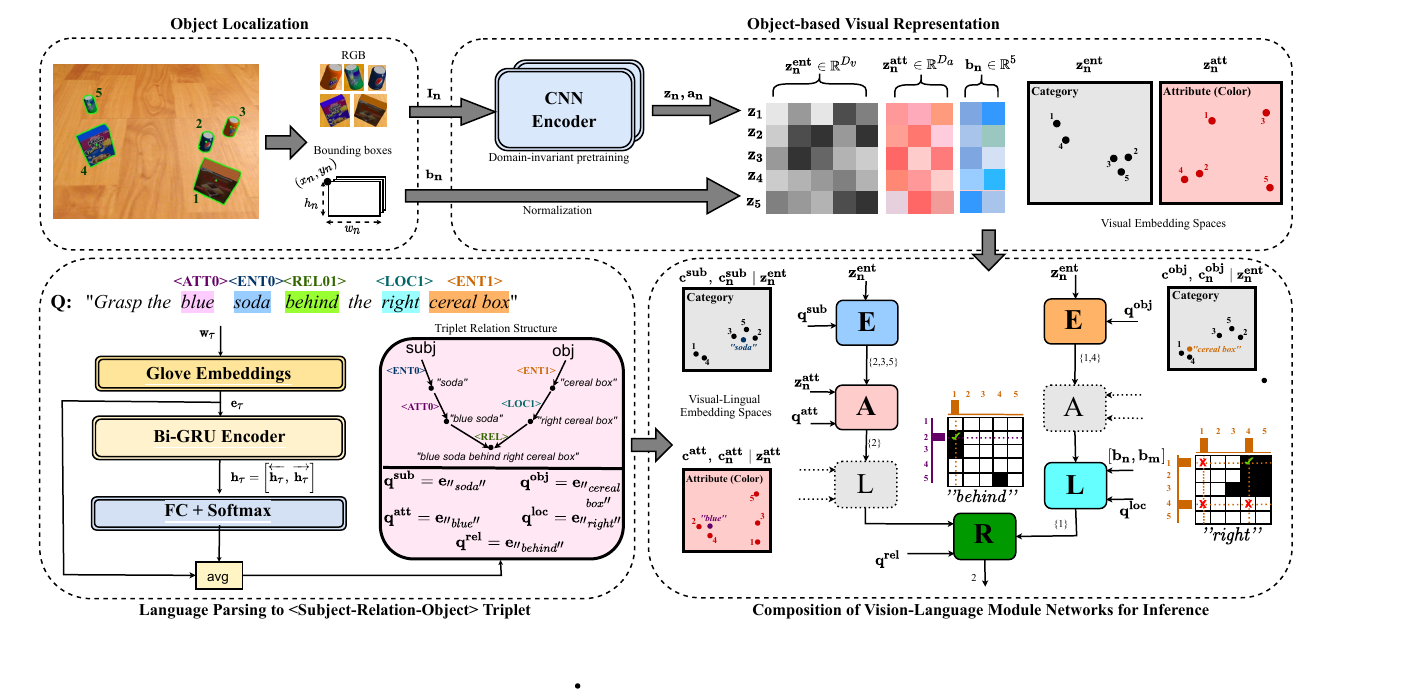}
    \caption{ A schematic of the proposed framework, illustrated with a running example for the given image and the query: \textit{``Grasp the blue soda behind the right cereal box''}. First, a localization module extracts object crops. Each crop is passed through a pair of CNN encoders (one for entity and one for attributes) and is concatenated with its bounding box features to constitute an object-based representation. For the language input, a word tagger assigns roles on each word in the sentence which indicates their grammatical role in the phrase. Role tags are replaced by (averaged) word embeddings and the triplet structured resolving the relation is mapped to an assembly of modules for inference. }
    \vspace{-2mm}
    \label{fig:system}
\end{figure*}
Visual grounding has been a central theme in recent deep learning literature, with an array of methods originating from different communities and for different applications. 
Most methods employ a two-step training strategy, with the first step being the pre-training of an object detector, therefore suffering from being bound to a pre-defined object list. 
More recent works address this with representation learning techniques, such as zero/few-shot grounding \citep{zsg, 1stage}, self-supervised \citep{visual-bert, vilbert} or multi-task \citep{12in1} pre-training. 
However, associated benchmark datasets struggle to cover generic, domain-agnostic content and thus rely on massive scale to be transferable effectively. 
Moreover, their transfer performance often suffers due to the domain discrepancy between the ideally pre-processed images of the training benchmarks and the noisy real-time sensory images. 
The alternative would be to design a domain-specific dataset directed towards HRI applicability (as in \citet{hatori}), something which comes at the great cost of having to manually collect data and provide linguistic annotations depending on the environment/scenarios that the agent is operating. 
In contrary, robotic manipulation and perception have been drastically benefited by the \textit{Sim-to-Real Transfer} paradigm, which offers a platform for data-efficient, domain-specific solutions that omit the need for manual annotation by utilizing tools to simulate the real world. 

In this paper, we propose a novel deep learning approach for Sim-to-Real transfer of visual grounding. The model breaks down the overall task to grounding of entities, attributes, and spatial relations separately. 
We develop a synthetic domain of household objects in a simulation environment and extract scene graph annotations that provide pair-wise relations for objects and their properties. 
We use this dataset to train our modules in a fully decoupled fashion and integrate it with a domain adaptation approach for visual recognition in order to explore its Sim-to-Real transfer potential. In summary, our key contributions are threefold:
\begin{itemize}
    \item We propose a new deep learning architecture that combines two modular formulations for decoupled visual grounding of entities, attributes, and spatial relations, while enabling strong supervision from pair-wise relation information.  
    \item We generate a synthetic dataset of household objects in tabletop scenes, paired with a small-scale real counterpart, aimed for evaluating Sim-to-Real transfer for visual recognition and grounding. Moreover, we establish a baseline for these tasks using our model combined with a popular domain adaptation technique.
    \item We perform extensive sets of experiments on synthetic and real datasets to show the benefits of our model in terms of data-efficiency and robustness to input query complexity. Furthermore, we integrate our system with a simulated and a real robot to use it as a proxy for visual grounding in an HRI scenario.
\end{itemize}

\section{Related work}
Although an in-depth review is beyond the scope of this work, we discuss a few recent efforts in \textit{visual grounding}, \textit{module networks} and \textit{robotics-related synthetic domains}.

\paragraph{Visual grounding:} Also called \textit{referring expression comprehension}, 
visual grounding  concerns the task of localizing a target region in an image according to a natural language query.  They key objective is that the model interprets linguistic information to reason about the scene, including entity properties and relations.
Many different approaches have been developed for visual grounding that can be organized into four main categories, including: (\textit{i}) joint-embedding models, which use either CNN-LSTMs or attention mechanisms to rank regions according to the conditional probability given the expression \citep{traditional1, traditional2, baselineMLP, baselineAttn}; (\textit{ii}) approaches that use an external parser to retrieve the order of referential reasoning exploiting linguistic cues \citep{nmtree, extparser}; (\textit{iii}) graph-based methods, which represent objects as nodes and their relations as edges \citep{graph1, graph2, graph3}, attempting to model complex relationships by updating neighbouring information with message passing; and (\textit{iv}) transformer-based pre-trained models, that follow the success of pre-training large transformer-based models in NLP for vision-and-language data, either in a self-supervised \citep{visual-bert, vilbert, UNITER} or a multi-task \citep{12in1} formulation. Most recently, transformer architectures have been employed for standalone visual grounding \citep{vgTRM}. 

\paragraph{Module networks} Architectures based on neural module networks \citep{NMN} have been studied for vision-and-language learning tasks, both in visual grounding \citep{CMN, Mattnet} and in VQA context \citep{modules2020, ns-cl, ns-vqa}. 
Similar to our approach, module networks break the overall architecture into separate modules, each responsible for grounding a specific concept in the input query. We base our approach on CMN \citep{CMN}, which assumes a triplet subject-relation-object structure in the input query and ground entities and relations separately by aggregating pair-wise matching scores. 
To deal with absolute relations that break the assumed triplet structure (e.g. \textit{"The right bowl"} is equivalent to \textit{"The bowl right from the other bowls"}), \citet{Mattnet} introduces a location module that aims at grounding relations between objects of the same category.
Moreover, they aid entity grounding by treating attribute prediction as a separate branch in the entity module.
In this work we aim to combine ideas from both methods to develop a model that deals with attributes and absolute location while maintaining the triplet assumption, which provides a strong prior for generalizing to more complex queries.
Both mentioned models are trained end-to-end in the weakly supervised setting, using only the ground truth bounding box as the supervisory signal, as pair-wise relation annotations are rarely provided in benchmark datasets. 
In contrary, we further exploit rich supervision from a synthetic dataset and train all modules in a decoupled fashion, which as we show in section~\ref{sim2real} allows for data-efficient adaptation to real scenes, while entertaining high interpretability.

\paragraph{Robotics-related synthetic domains} Pre-training in synthetic datasets is a broadly adapted practise for robotic manipulation \citep{manip,drl_survey,kadian2020sim2real}, planning \citep{nav, nav2} and 3D vision tasks \citep{3d1, 3d2}. For visiolinguistic data, the field is much drier, with barely any benchmarks for vision-and-language tasks in RGB-D data.
An exception is the VQA benchmark SHOP-VRB \citep{SHOP-VRB} which models a kitchen environment of $20$ distinct objects, with rich attribute annotations (e.g weight, mobility etc.).
The development of the benchmark is based on \textit{CLEVR} \citep{clevr}, a synthetic dataset of abstract object shapes in arbitrary spatial arrangements. The dataset is equipped with question-answer-functional program annotations, used mainly to diagnose the visual reasoning abilities of established models
Similarly, we build on top of the CLEVR annotation engine to automatically sample visual grounding queries from scene graph representations that we generate in a simulation environment.
\label{sota}
\section{Methodology}
\label{methods}

\subsection{Problem Formulation}
We formulate the visual grounding task in the retrieval setting. 
Given the RGB frame of the input scene, represented as a set of $N$ extracted \textit{Region of Interest (ROI)} proposals, $\mathbf{o}_n$, and a natural language query, $Q$, that refers to a specific object depicted, the goal is to find the $n^* = \texttt{argmax}_{n} ~ S(\mathbf{o}_n, Q)$, where $n \in \{1,\dots,N\}$ and $S_n:=S(\mathbf{o}_n, Q)$ is a similarity function between any ROI and the query.

We adopt the subject-relation-object triplet composition method of \citet{CMN}, but adapt our modules to also include absolute location queries, such as \citet{Mattnet}.
More specifically, we formulate the similarity function as $S_n:=R(S_n^{sub}, S_n^{obj})$, where:
\begin{equation}
\label{equ1}
    R(S_n^{sub}, S_n^{obj})=\max\limits_{m \in \{1,\dots,N\}} ~ (S_n^{sub} + S_{nm}^{rel} + S_{m}^{obj})
\end{equation}
This function estimates the similarity between each ROI and the query (e.g. \textit{``the bowl next to the bottle''}), by aggregating unitary matching scores between the ROIs and isolated words / phrases that correspond to the two syntactic roles, i.e. subject $\mathbf{q}^{sub}$ (e.g. \textit{``bowl'')} and object $\mathbf{q}^{sub}$ (e.g. \textit{``bottle''}), with pair-wise scores between the ROIs and a relation word $S_{nm}^{rel} = S^{rel}(\mathbf{o}_n, \mathbf{o}_m, \mathbf{q}^{rel})$ (e.g. \textit{``next to'')}.
The unitary scores for each ROI are computed after composing two vision-language matching modules, one for \textit{entity} (\texttt{E}) and one for \textit{attribute} (\texttt{A}) matching, as well as potentially one spatial-language for \textit{location} (\texttt{L}) matching, such that for $k=\{sub,obj\}$:
\begin{equation}
S_n^{k} = \texttt{E}(\mathbf{o}_n, \mathbf{q}_{k}^{ent}) \circ \texttt{A}(\mathbf{o}_n, \mathbf{q}_{k}^{att}) \circ \texttt{L}(\mathbf{o}_n, \mathbf{q}_{k}^{rel})
\label{equ2}
\end{equation}

This formulation combines benefits from both related modular approaches, on the one hand enabling training with \textit{strong supervision} through pair-wise relation annotations for each scene (as discussed in \citet{CMN}), while at the same time incorporating relation cases of absolute location (e.g. \textit{``the left bowl''}) that break the assumed triplet structure.
Similar to \citep{Mattnet}, we treat visual attribute grounding as a standalone module, disentangled from semantic grounding (i.e. the category of the object).
Another departure from the previous methods is the replacement of the soft language parser with a standalone hard parser, trained solely in synthetic data. 
We highlight that this is not equivalent to methods utilizing external parsers \citep{external1, NMN}, as even though the parser is not trained end-to-end with the rest of the architecture, it is yet optimized for the specific domain and its representations are used for training the three matching modules.


\subsection{Object-Based Representation}
In our visual grounding setup, all appearing objects are first localized, and visual features for each ROI crop are then extracted. Object localization is most popularly achieved by integrating advanced methods for detection (e.g. Mask-RCNN \citep{he2017mask}) or clustering (e.g. UCN \citep{xiang2020learning}).
However, we exploit our robotic setup and localize objects directly in 3D space.
In particular, we first remove the planar point cloud (table), cluster the remaining points~\citep{rusu20113d}, and finally calculate a local reference frame by applying PCA on the normalized covariance matrix of each cluster~\citep{kasaei2018towards}.
The 3D coordinates for each object are translated to the RGB camera reference frame and the corresponding ROI, $\mathbf{I}_n$, is cropped from the input image using the translated 2D bounding box coordinates.

We treat visual attributes of appearing entities (such as color, material, etc.) as a means of disambiguating objects of the same category (e.g. \textit{``red bowl''}, \textit{``plastic cup''}) and attempt to ground them independently of the object categories~\citep{attributes3, attributes4}.  
To obtain object-based representations $\mathbf{o}_n$ for each ROI, we pass the image crop $\mathbf{I}_n$ to two visual feature extractors, one associated with the entity $F^{E}$ and one with the attribute $F^{A}$ modules, such that:
\begin{equation}
\mathbf{o}_n = \left [ \mathbf{z}_n^{ent}, ~ \mathbf{z}_n^{att}, ~ \mathbf{b}_n, \right ]
\end{equation}
\begin{equation}
\mathbf{z}_n^{ent} =  F^{E}(\mathbf{I}_n ;\mathbf{\Theta}_{E}), ~ ~ ~ \mathbf{z}_n^{att} = F^{A}(\mathbf{I}_n;\mathbf{\Theta}_{A})
\end{equation}
with $\left [ , \right ]$ denoting the concatenation operation and $\mathbf{\Theta}_E, \mathbf{\Theta}_{A}$ represent trainable parameters of the two CNN encoders $F^{E}:\mathbb{R}^{H \times W \times 3} \rightarrow \mathbb{R}^{D_v}$ and $F^{A}:\mathbb{R}^{H \times W \times 3} \rightarrow \mathbb{R}^{D_a}$.
For entities, we start from a pre-trained CNN checkpoint and keep until the penultimate layer of the network, 
For attributes, an earlier (second or third) layer is more appropriate, as lower-level layers provide more abstract features (e.g. color, shape etc.) that are more useful for attribute matching \citep{Mattnet}. Alternative representation methods can also be utilized (e.g. color histograms).
We further inject all spatial information lost due to cropping ROIs by concatenating the visual representations with normalized bounding box features, $\mathbf{b}_n \in \mathbb{R}^5$, where:
\begin{equation}
\mathbf{b}_n = \begin{pmatrix}
\frac{x_n}{W} & \frac{y_n}{H} & \frac{x_n+w_n}{W} & \frac{y_n+h_n}{H} & \frac{w_n \cdot h_n}{W \cdot H}
\end{pmatrix}^T
\end{equation}

We adopt such object-based visual representations to aid their Sim-to-Real adaptation, as training in ROI crops, instead of the entire RGB frame, acts as a first step to reduce the domain gap between the source and target domains.

\subsection{Language Parser}
\label{lang}
We parse the input query by mapping each word $\mathbf{w}_{1:T}$ in a phrase of length $T$ to a tag $\mathbf{u}_{1:T}$, which serves as an indicator for the word's role in the sentence (incl. entity, relation, location and attribute).
We utilize Glove embeddings \citep{glove} to embed the input words and a bi-directional GRU \citep{gru} of hidden size $D_h$ to contextualize embeddings within the phrase. 
The hidden state at each step $\tau=1, \dots, T$ is given by the concatenation of the hidden states for the two directions, such that:
\begin{equation}
    \mathbf{e}_{\tau} = \texttt{GloVeEmbedding}(\mathbf{w}_{\tau})
\end{equation}
\begin{equation}
    \mathbf{h}_{\tau} = \left [\overleftarrow{\mathbf{h}_{\tau}}, ~ \overrightarrow{\mathbf{h}_{\tau}} \right ]
\end{equation} 
\begin{equation}\overleftarrow{\mathbf{h}_{\tau}}=\texttt{GRU}(\mathbf{e}_    {\tau}, ~ \overleftarrow{\mathbf{h}_{\tau-1}}), ~ ~ ~     \overrightarrow{\mathbf{h}_{\tau}}=\texttt{GRU}(\mathbf{e}_{\tau}, ~ \overrightarrow{\mathbf{h}_{\tau-1}})
\end{equation}

We extend our tag set with two additional concepts, a \textit{mask} tag that tells our model to ignore the specific word (e.g. \textit{``The''} in \textit{``The red bowl''}) and an \textit{abstraction} tag that is used for words that generalize the concept of an object (e.g. \textit{``item''} in \textit{``The item next to the bowl''}). The role tag is followed by an integer $0-2$ that indicates which attribute/location should be grouped with which entity phrases.
The resulting tag vocabulary enumerates a total of $D_u=3\times6=18$ tags.
Unlike soft parsing methods, this allows the model to extend the structure recursively up to two-hop relation queries (e.g. \textit{``The bowl behind the item that is left from the bottle''}).
Overall, we tag each word by applying a linear projection $\mathbf{\Theta}_u \in \mathbb{R}^{2 \cdot D_h \times D_u}$ on of the GRU output, followed by a softmax activation: $\mathbf{u}_{\tau} = \texttt{softmax} \left ( \mathbf{\Theta}_u^T \cdot \mathbf{h_{\tau}} \right )$.

We recover the desired triplet structure by composing all $\mu \in \{ent, att, loc\}$ tags per unique group index $k=0,\dots,2$ and replace the role tags with (averaged) word embeddings of the corresponding words / phrases:
\begin{equation}
    \mathbf{q}^{\mu}_k=\texttt{avg}_{\tau} \{\mathbf{e_{\tau}} \mid \mathbf{u_{\tau}}=[\texttt{TagVocab}(\mu), k] \}
\end{equation}
\begin{equation}
    Q_k =\{\mathbf{q}^{ent}_k, \mathbf{q}^{att}_k, \mathbf{q}^{loc}_k  \}
\end{equation}
For standard single-hop relation queries $k \in \{0,1\}$, the two identified groups will result in the $\{\mathbf{q}^{ent}_k, \mathbf{q}^{att}_k, \mathbf{q}^{loc}_k  \}$ embeddings for the subject/object module evaluations appearing in equation~\ref{equ2}.
For a detailed example parse, see bottom-left of Fig.~\ref{fig:system}.

\subsection{Matching Modules}
\label{modules}
We implement a two-layer MLP to embed the object-based and the extracted query embeddings into the same space, where we compute a unitary matching score.
Specifically, for unitary modules $E, A$, given the visual representation $\mathbf{z_n^{\mu}}$ and a word embedding $\mathbf{q^{\mu}}$, the final matching score will be given by:
\begin{equation}
    \mathbf{c}_n^{\mu} = (\mathbf{\Theta}_1^{\mu} \cdot \mathbf{z}_n^{\mu}) \odot \mathbf{q^{\mu}}, ~ ~ \mu=\{ent,att\}
\end{equation}
\begin{equation}
\texttt{E}(\mathbf{q}^{ent})=\mathbf{\Theta}_2^{ent} \cdot \frac{\mathbf{c}_n^{ent}}{\left \| \mathbf{c}_n^{ent} \right \|_2}, ~ ~ \texttt{A}(\mathbf{q}^{att})=\mathbf{\Theta}_2^{att} \cdot \frac{\mathbf{c}_n^{att}}{\left \| \mathbf{c}_n^{att} \right \|_2}
\end{equation}
\begin{equation}
\label{equ13}
    S^k_n = \texttt{E}( \mathbf{q}^{ent} ) \odot \texttt{A} ( \mathbf{q}^{att})
\end{equation}
where $\odot$ denotes element-wise multiplication and $\mathbf{\Theta}_1^{ent} \in \mathbb{R}^{D_v \times D_jv}, \mathbf{\Theta}_2^{ent} \in \mathbb{R}^{D_jv \times 1}, \mathbf{\Theta}_1^{att} \in \mathbb{R}^{D_a \times D_jv}, \mathbf{\Theta}_2^{att} \in \mathbb{R}^{D_jv \times 1}$  trainable matrices of the the entity and attribute modules respectively. 
We formulate the location (\texttt{L}) module on a similar matching network $\texttt{B}$, which is used for spatial-language matching. 
Given two objects $n,m \in \{1,\dots,N\}$, we concatenate their location features $b_{nm} = [b_n, ~ b_m]$ and use the same architecture as above for getting one score per object pair:
\begin{equation}
    \texttt{B}(\mathbf{q}^{spt})=\mathbf{\Theta}_2 \cdot \frac{(\mathbf{\Theta_1} \cdot \mathbf{b}_{nm}) \odot \mathbf{q}^{spt}}{\left \|  (\mathbf{\Theta}_1 \cdot \mathbf{b}_{nm}) \odot \mathbf{q}^{spt}\right \|_2}
\end{equation}

with $\mathbf{\Theta}_1^{spt} \in \mathbb{R}^{10 \times D_js} , \mathbf{\Theta}_2^{spt} \in \mathbb{R}^{D_js \times 1}$ trainable parameters of \texttt{B}. The resulting $N \times N$ matrix gives highest scores for objects that respect the spatial relation encapsulated in  $\mathbf{q^{spt}}$.
In order to locate the input object, the aggregated entity-attribute scores of equation~\ref{equ13} are given as input to the location module and the final output is given according to equation~\ref{equ2}:
\begin{equation}
\texttt{L}( \mathbf{q}^{spt})= \texttt{softmax} ~ \left ( \sum_{m}^{}S_n^k \cdot S_m^k \cdot \texttt{B}( \mathbf{q}^{spt}) \right )
\end{equation}

The spatial-language module $\texttt{B}$ provides the pair-wise scores $S_{nm}^{rel}$ that combined with the aggregated unitary scores for subject/object are used to express the training objective of  equation~\ref{equ1}.
A running example of module composition for inference is shown on the bottom right of Fig.~\ref{fig:system}.

\section{Experimental Results}
To assess the performance of the proposed approach, we perform three rounds of experiments, including: (\textit{i}) evaluation of visual grounding performance in a synthetic dataset generated in simulation, (\textit{ii}) evaluation of the adaptability of our method in real-world scenes after few-shot fine-tuning of the visual representations using two RGB-D datasets and (\textit{iii}) evaluation of the applicability of our method in a HRI scenario for human-aided robotic grasping in both simulation and a real robot.

\subsection{Datasets}
We focus our experiments on tabletop domains where objects are in arbitrary spatial arrangements. Objects are annotated both in category (e.g. bowl, soda can, etc.) as well as instance-level (red/white bowl, coca-cola/pepsi etc.) and the most dominant color of the object is used as the only attribute for visual disambiguation. For disambiguating through spatial relations, we include the following spatial concepts: \{\textit{``left'', ``right'', ``further'', ``closer'', ``behind'', ``front'', ``bigger'', ``smaller'', ``next to''}\}. As in most natural settings, humans often refer to objects using various words other than the predefined category label (e.g.. \textit{``Bring me the coca-cola''} is more natural than \textit{``Bring me the red soda can''}), we also include annotations for appropriate items according to their brand, variety or flavour (e.g. \textit{``Coca-Cola''} vs \textit{``Pepsi''}, \textit{``Chex''} vs \textit{``Crunch''}, \textit{``strawberry juice''} vs. \textit{``mango juice''}).

\subsubsection{\textbf{Synthetic Scenes}}
\begin{wrapfigure}{r}{0.35\textwidth}
    \vspace{-3mm}
    \centering
    \includegraphics[width=0.35\textwidth]{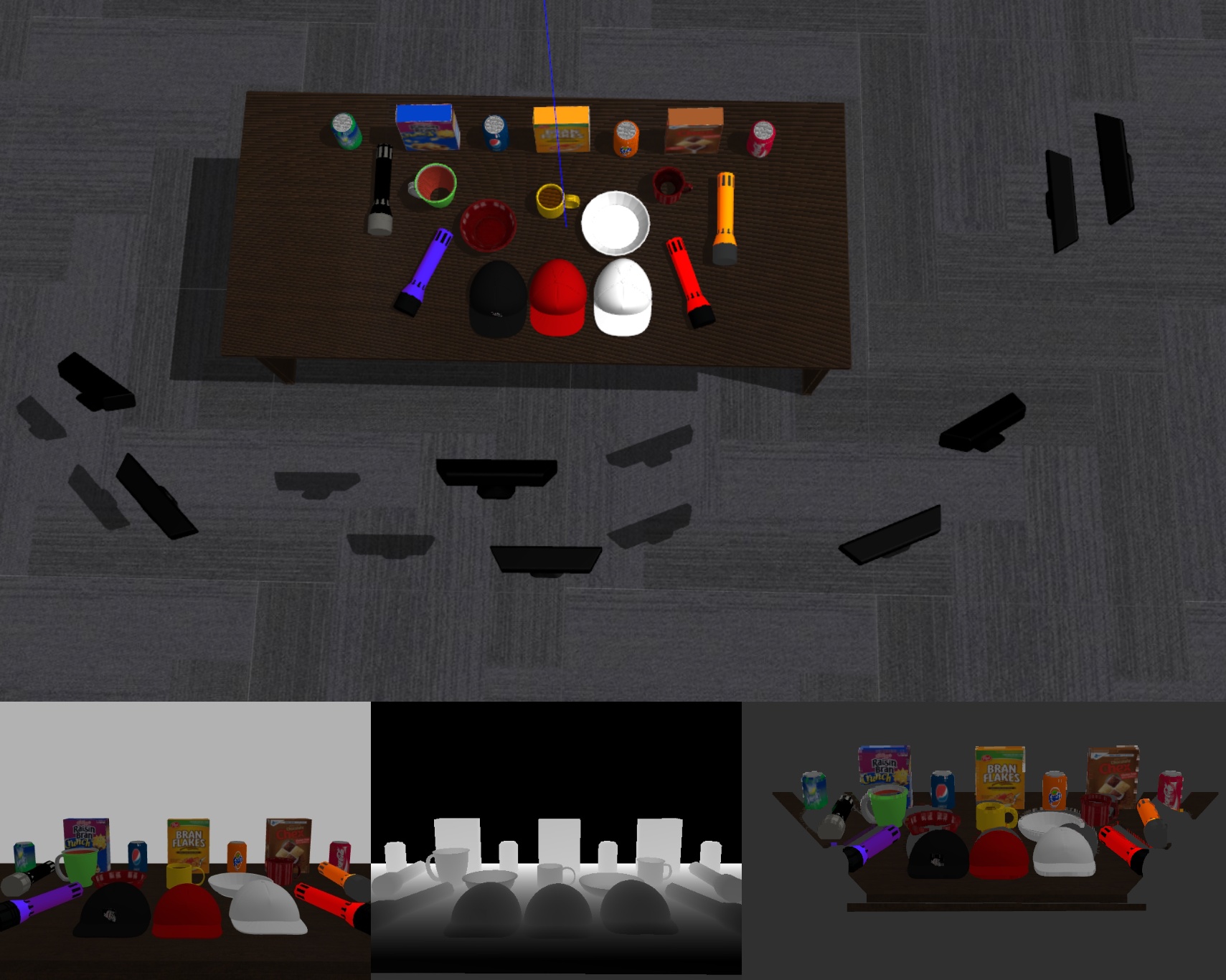}
    \caption{Synthetic recreation of a tabletop scene in Gazebo. The object catalogue (top), an example front view RGB (bottom-left) and depth (bottom-middle) pair, as well as the corresponding point cloud visualized in RViz (bottom-right).}
    \label{fig:gazebo}
    \vspace{-2mm}
\end{wrapfigure}
We collect a set of $32$ object models from available resources and organize them into $8$ object categories, with a total of $10$ different colors. We render synthetic scenes in the Gazebo environment \citep{Gazebo} and develop tools for automated scene annotation, including class labels for categories and color, bounding boxes and segmentation masks for localization, as well as pair-wise spatial relations for all objects in the scene. 
Each scene contains $5$ to $10$ objects and all parsed scene information including object attributes and their relations are gathered in symbolic scene graph representations. We generate a total of $8$k scenes for the training and $2$k for the development set and multiple test, splits aiming to evaluate different components of the proposed method. 

In order to generate natural language queries from the gathered scene graphs, we develop on top of the CLEVR \citep{clevr} data generation engine with the following adaptations: a) we treat object category as the main object identifier (instead of shape), b) we exclude material attributes and treat size as a spatial concept, c) we sometimes replace category with special instance-level words if applicable, d) we extend the vocabulary of spatial concepts to the ones aforementioned and e) we add absolute location annotations phrased as attributes. We use the $\{0,1,2\}$-hop question templates after changing the text from the QA to the desired grounding format and implement constraints to control the ambiguity of the generated queries.

\subsubsection{\textbf{Real Scenes}}
We evaluate the transfer performance of our model in real scenes from two RGB-D datasets, the publicly available Washington RGB-D scenes v1 dataset \citep{washv1}, and a manually collected dataset (Custom).

\paragraph{{\textbf{Washington Scene V1}}} 

We evaluate our approach on the Washington RGB-D scenes dataset \citep{dataset}, comprised of $8$ different scene setups (three tables, three desks, a kitchen and a meeting room table). The dataset contains six object categories from a total of $19$ different object instances, usually discriminated by visual cues (e.g. yellow vs. blue flashlight), although also containing object instances of the same category and color (e.g. two white bowls).

\paragraph{\textbf{Custom}} 

The object catalogue is a subset of the objects included in Gazebo, but include novel attribute combinations (e.g purple soda can, orange Pringles etc.), containing a total of $17$ object instances in $5$ categories (soda can, Pringles, juice box, book and milk box). We place each object in the table in uniform positions and orientations and collect up to $20$ samples per object instance, aimed as a fine-tuning resource for the visual representation backbone. We further record $206$ test scenes containing $5-15$ arbitrarily chosen objects placed randomly across the workspace and annotate them for location and visual attributes manually. We then follow the same process as in simulated data to construct annotated scene graphs as well as sample natural language queries.

\subsection{Visual Grounding Evaluation in Simulation}
In this section we explore the benefits of the proposed approach for visual grounding in the synthetic domain.  
We compare our method with two end-to-end baselines, namely: \textit{(i)} \textbf{feature-fusion}, following the CNN-LSTM + MLP architecture of \citet{baselineMLP}, and \textit{(ii)} \textbf{soft-parser-based} modular networks, based on CMN \citet{CMN} and MattNet \citet{Mattnet}. 
We highlight that even though modular, the second baseline's networks require training of all modules in an end-to-end fashion, unlike our decoupled approach.
For architectural and training details of our implementations for this experiment, see Appendix~\ref{AppendixB}. 

In order to diagnose the effectiveness of adding explicit modules and hard parsing, we ablate the modules of each compared method and report results broken down by type of disambiguation used in the query. 
In particular, we include: (\textit{i}) \textbf{Category (Cat)}, where there is no present ambiguity and the category of the object is used (e.g. \textit{``bowl''}), (\textit{ii}) \textbf{Color (Col)}, referring to the color of an object (e.g. \textit{``blue bowl''}) (\textit{iii}) \textbf{Location (Loc)}, where an absolute spatial relation between objects of the same category is used (e.g. \textit{``rightmost bowl''}), (\textit{iv}) \textbf{1-Hop}, where there is a single subject-relation-object triplet without further attributes/locations (e.g. \textit{``bowl behind flashlight''}), as well as (\textit{v}) \textbf{2-Hop}, where one triplet acts as the object phrase for another triplet and attributes/locations might be present within all phrases (e.g. \textit{``blue bowl behind the flashlight that is closer than the red mug''}). 
We evaluate our models in the development set and select the best performing model using early stopping.
We then report the performance of the selected models in various test splits, using top-1 accuracy of the predicted ROI index as the evaluation metric.

\begin{wraptable}{r}{0.55\textwidth}
    \vspace{-3mm}
    \centering
    \resizebox{\linewidth}{!}{
    \begin{tabular}{lccccccc}
     \textbf{Model} & \textbf{Cat} & \textbf{Col} & \textbf{Loc} & \textbf{1- Hop} & \textbf{\textbf{2-Hop}} & \textbf{Average} \\
     \toprule
     \emph{(E) + GRU + MLP} & $98.7$ & $67.8$ & $51.4$ & $59.1$ & $56.4$ & $66.7$ \\
     \emph{(E+A) + GRU + MLP}& $98.5$ & $95.4$ & $51.3$ & $58.4$ & $54.9$ & $71.7$ \\
     \emph{(E+A+L) + GRU + MLP}& $98.1$ & $95.4$ & $65.5$ & $60.0$ & $55.6$ & $74.0$ \\
	\toprule
     \emph{ (E) + SoftParser}   & $\mathbf{98.8}$ & $72.9$ & $50.7$ & $66.3$ & $51.4$ & $68.0$ \\
     \emph{(E+A) + SoftParser}   & $\mathbf{98.8}$ & $95.5$ & $50.7$ & $72.4$ & $52.7$ & $74.0$ \\
    \emph{(E+A+L) + SoftParser}   & $\mathbf{98.8}$ & $95.5$ & $\mathbf{98.7}$ & $96.3$ & $50.2$ & $87.9$ \\
     \toprule
     \emph{(E+A+L) + HardParser} & $98.4$ & $\mathbf{95.7}$ & $98.1$ & $\mathbf{96.9}$ & $\mathbf{96.3}$ & $\mathbf{97.0}$ \\
    \end{tabular}}
    \caption{Visual grounding accuracy (\%) evaluated for the test split of the synthetic data, organized by type of reference. Baselines include CNN-LSTM feature fusion (1-3), end-to-end modular using a soft language parser (4-6) and modular using hard language parsing (7). Compared to fusion-based, modular approaches solve relation queries by treating spatial relation comprehension separately. Without incorporation of explicit modules (i.e. for \textit{attributes (A)} and \textit{location (L)}), all baselines struggle to perform in corresponding splits. Disentangling language parsing from assembling modules for inference allows generalization to multiple relation steps entailed in the input query \textit{(see 2-Hop}).  
    }
    \label{tab:res1}
\end{wraptable}

The results are summarized in Table~\ref{tab:res1}. 
A first observation is that the feature-fusion baselines struggle to comprehend spatial relations, even when concatenating location features in the objects representations. 
This is due to lack of contextualization between locations of different objects.
Second, the explicit incorporation of attribute features, either just concatenating in the holistic or as a separate module for the modular approaches, results in very high accuracy for color queries.
This result positively reinforces using independent representations for visual category and attributes, as very commonly high-level pre-trained representations have "dropped" abstract features (such as color, shape) in favour of more category-specific ones (see $67.8\%$ over $95.0+\%$ in column 2 of the table).
A similar result is observed regarding absolute spatial relations, where using a separate location module for absolute relations resolves the grammatical discrepancy of relevant queries from the standard triplet structure.
This leads to big accuracy gap in the locations split.
Finally, we observe that even though end-to-end approaches successfully ground single-hop queries, they consistently cannot generalize to two-hop relations. 
We attribute this result to the fact that these approaches use a single relation concept assigned, resulting in both relation words present in the sentence aggregated and the entire second subject-relation-object triplet treated as single object.
On the other hand, our decoupled formalism naturally generalizes to higher-order hops, as each hop is predicted uniquely in the word tagging step, resulting in a huge performance gap with the other baselines ($96.3\%$ vs. $<56.4\%$.)

\subsection{Sim-To-Real Adaptation Evaluation}
\label{sim2real}
In this subsection we evaluate the potential of our method for few-shot adaptation to real scenes. 
The decoupled nature of our model allows us to battle the gap between sim and real solely on the visual domain, using a domain adaptation method for object recognition as a representation technique and directly transferring the rest of the architecture without any further training.
We highlight that this is a unique benefit of our approach compared to soft-parser-based baselines, as due to their end-to-end nature they also need additional text data to fine-tune the soft-parser as part of the architecture (see Appendix~\ref{AppendixC}) for a discussion and data-efficiency comparisons).

In this round of experiments, we pre-train our CNN encoder paired with a classification head, in order to align the representations of objects in the two domains, using only a few real labeled images.
We replace \textit{MobilenetV2} features with the pre-trained domain-invariant representations, and then fine-tune the entity/attribute modules of our hard-parser-based model in synthetic data.
We use the DIRL \citep{dirl} domain adaptation method  and compare it with four baselines, namely: (\textit{i}) using only synthetic, (\textit{ii}) using only the few real data, (\textit{iii}) training in both sim and real datasets, and (\textit{iv}) the unsupervised method DANN \citep{dann}.
Both mentioned techniques employ adversarial training to align the distributions between the two domains, with DIRL exploiting a few labeled examples to further model the conditional (domain discrimination per class) together with the marginal (unsupervised domain discrimination) distribution. 
We refer the reader to the original paper for more theoretical insight on domain adversarial training.
For implementation details and visualizations of the training progress, see Appendix~\ref{AppendixC}.

\begin{wraptable}{r}{0.5\textwidth}
    \vspace{-3mm}
    \centering
    \resizebox{\linewidth}{!}{
    \begin{tabular}{lcccccc}
    \multirow{2}{1.5cm}{\textbf{Model}} &
    \multicolumn{3}{c}{\textbf{Washington v.1}} &
    \multicolumn{3}{c}{\textbf{Custom}} \\
       & Rec & Ent/Att & VG & Rec & Ent/Att & VG \\
     \toprule
      Sim only  & 39.9 & 38.5 & 29.1 & 22.4 & 20.0 & 12.9 \\
      Real only  & 61.9 & 58.4 & 55.0 & 78.2 & 74.9  & 71.3 \\
      Sim+Real  & 58.6 & 56.2 & 52.1 & 74.8 & 72.3  & 68.8\\
     \hline 
     DANN  & 73.4 & 72.7 & 66.9 & 66.8 & 65.3 & 60.2\\
      DIRL & 88.6 & 86.2 & 81.0 & 86.1 & \textbf{85.9} & 80.1 \\
      DIRL (+augm.+sel.)  & \textbf{90.8} & \textbf{88.0} & \textbf{81.3} & \textbf{87.1} & 85.7 & \textbf{80.5} \\
    \end{tabular}}
    \caption{Evaluation results for sim-to-real adaptation of visual representations for the test splits of both RGB-D datasets. Results are reported as top-1 predicted accuracy (\%) evaluated using adapted embeddings, including (from left to right): recognition accuracy \textit{(Rec)}, average of entity-word \textit{(Ent)} and attribute-word \textit{(Att)} matching accuracy, as well as overall performance for visual grounding \textit{(VG)}. Final row indicates manually selecting the labeled subset of target images and extending it with random cropping - illumination augmentations.}
    \label{tab:res2}
\end{wraptable}

We report object recognition accuracy of the DIRL classifier after pre-training, the accuracy of our entity and attribute modules after training using the adapted representations and the overall visual grounding accuracy for synthetic queries generated for the test splits of the two target datasets.
The last experiment includes queries from all the ambiguity types of Table~\ref{tab:res1}.
Our results are presented in Table~\ref{tab:res2}. 
We observe that DIRL outperforms other baselines in both datasets and overall provides robust representations for extending to the visual grounding task.
Performance for matching and grounding closely follows that of visual recognition in all baselines, suggesting that the only bottleneck in performance is the quality of the visual representations.
Overall, the system demonstrates high data-efficiency for adaptation to natural images. Notably, the system achieves $81.3\%$ accuracy on Washington v1 and $80.5\%$ on our custom dataset, while only being exposed to real supervision from $20$ labeled images per object category.
Most importantly, no language data for the target datasets are required in order to transfer the grounding system.
We believe that this result strongly suggests in favour of utilizing modeling and learning in synthetic domains for vision-and-language tasks, paired with domain adaptation techniques for visual data only.

\subsection{Integration with a Robot Framework for Human-Aided Object Picking}
We integrated the proposed visual grounding pipeline into the RACE cognitive robotic framework~\citep{kasaei2018towards} and evaluated the end-to-end system in a HRI scenario, where a non-expert human user instructs a robot to grasp an (ambiguous) object from a crowded scene. 
In particular, we use RACE as a world model that we call to obtain object locations and grasp proposals in real-time.
The user makes a query and after an object ROI is proposed, the most suitable grasp pose is sampled and the actuator is moved to the grasp position using an inverse kinematics solver. 
The robot then attempts to pick the object and return back to default position. 


\begin{wraptable}{r}{0.5\textwidth}
    \centering
    \resizebox{\linewidth}{!}{
    \begin{tabular}{lccccc}
     \textbf{Split} & \textbf{\#Trials} & \textbf{\#Fail.} & \textbf{\#Pars.F.}& \textbf{\#Perc.F.} &  \textbf{\#Grasp.F.}  \\
     \toprule 
    \emph{simple (Temp.)} & 90 & 11 (12.0\%) & 0 (0.0\%) & 1 (1.1\%)  & 10 (9.0\%)\\
    \emph{mixed (Temp.)} & 90 & 17 (21.1\%)  & 0 (0.0\%)  & 3 (3.3\%) & 16 (17.7\%) \\
    \toprule
    \emph{simple (Human)} & 45 & 7 (15.5\%) & 2 (4.4\%) & 1 (2.2\%) & 4 (8.8\%) \\
    \emph{mixed (Human)} & 45 & 14 (32.1\%) & 9 (20.0\%) & 2 (4.4\%) & 3 (6.6\%) \\
    \end{tabular}}
    \caption{Evaluating the system for coupling visual grounding with robotic grasping in synthetic scenes.}
    \label{tab:grasp}
\end{wraptable}
We generate random scenes in simulation and evaluate the end-to-end behavior of the system.
We report results separately for \textbf{simple} (i.e only color, only location, color+location or entity-only 1-hop) queries in images of $5-8$ objects and \textbf{mixed} (1- and 2-hop relations including colors and locations in arbitrary parts of the phrase) queries in images of $9-12$ objects.
All queries are given by a human supervisor in real-time after inspecting the simulated scenes.
To emulate potential discrepancies of our synthetic templates with natural language, we repeat the experiment for a smaller split where 3 human supervisors gave arbitrary queries without being aware of the template structure.
The interpretable nature of our model's inference allows us to diagnose the source of failure in case of errors, namely: a) \textbf{parsing} failure, where the input query is parsed incorrectly as a result of a mis-classification in the tagger network, b) \textbf{perception} failure, where the parse is correct but the matching modules gave incorrect scores, and c) \textbf{grasping} failure, where the system behaves correctly but the grasping behaviour fails.
Results are summarized in Table~\ref{tab:grasp}. 

We observe that in the template splits, the visual grounding system is robust to the grasping instructions across all trials, with the exception of a few perception failures.
Language parsing as well as spatial relation grounding completed all trials without any errors.
In the natural queries, we observe an increase in the parsing errors of the system, especially in the case of the mixed split. 
This is caused in almost all cases because of the use of an unknown spatial concept (i.e. \textit{"middle", "between"} etc), which leads in arbitrary spatial-language matching scores.
Perception failures remain very low ($<5\%$) and as in the previous split are always entity mismatches due to partial views of objects caused by occlusion.
\begin{figure*}[t]
    \centering
    \includegraphics[width=1\textwidth]{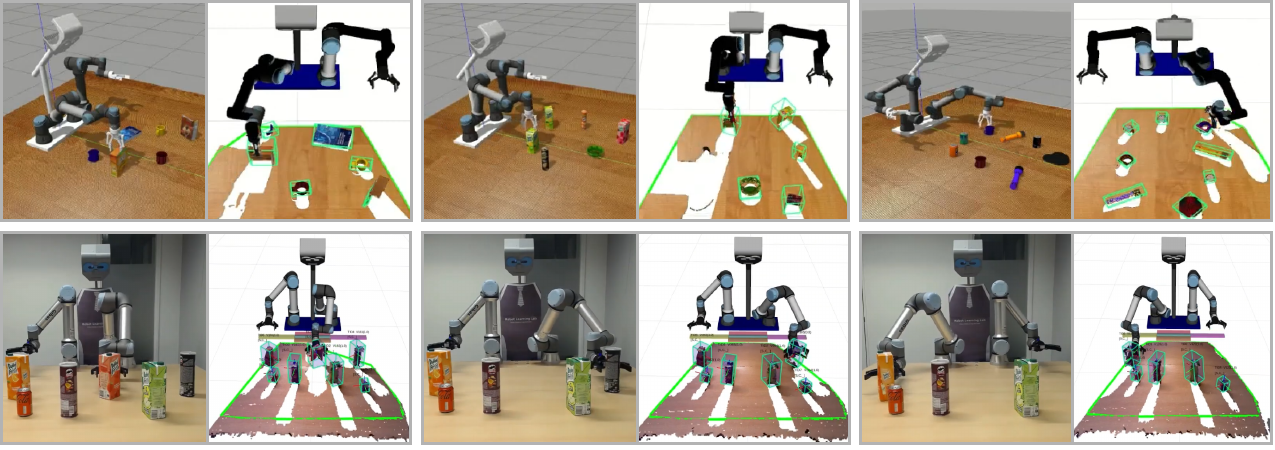}
    \caption{A sequence of snapshots capturing the setup of our robot framework in Gazebo (\textit{top)} and in a real-world environment (\textit{bottom)}. We generate a random scene and command the robot to grasp a specific item with a text instruction, using a disambiguation query. For our examples the queries are, from \textit{top-left} to \textit{right-bottom}: \textit{``cereal box behind the blue mug''}, \textit{``juice box next to the rightmost Pringles''}, \textit{``mug left from the soda can behind the flashlight''}, \textit{``pink juice box''}, \textit{``leftmost Pringles''}, \textit{``juice box right from the Pringles''}. In the snapshots we demonstrate the robot during the picking action (\textit{left)} and the localization results in RViz (\textit{``right''}).}
    \label{fig:snapshots}
\end{figure*}
The overall results showcase that the system can indeed be effectively utilized as a visual grounding utility for robotic grasping, with the main bottleneck being the grasping behaviour. 
Even in the case of human users not being aware of the robot's vocabulary, results suggest that the parsing module is general enough ($12.2\%$ on average for both splits) to capture the majority of different variations of language humans used to relate objects in the tabletop setups of the experiment.

Furthermore, we perform similar end-to-end grasping experiments with a real robot. Our setup is shown in Fig.~\ref{fig:snapshots}. 
For visual representation, we use the CNN encoder that we adapted for our Custom dataset. We place objects randomly within the workspace of the robot and query for different instances based on color, location and spatial relations between them. A video demonstrating the robot behaviour has been attached to the paper as supplementary material.

\section{Conclusion}
\label{Conclusion}
In this work, we present a novel modular deep learning framework for visual grounding in RGB-D domains. We focus on an ambiguity resolution HRI scenario, where the input query can always be decomposed in terms of grounding entities, attributes, and spatial relations between different objects. We develop a synthetic domain and tools for automatic annotation, which we use to train our modules independently. 
In particular, the controlled setup of simulation is used to generate elaborate scene graph representations, which are used to generate appropriate query expressions. For inference, we combine the modules by exploiting the grammatical priors of relations in natural language. Experimental results carried in simulation show that the proposed method achieves high accuracy in all versions of the ambiguity resolution task. 
Additionally, we show that, compared to other modular baselines, our method is robust to input query complexity, while the decoupled formulation of the modules enables high interpretability at test time.
We also assess the sim-to-real adaptation of the proposed approach on two real RGB-D scene datasets. 
Results showed that our method allows for data-efficient adaptation to real-scenes using only few-shot visual supervision. 
We finally integrate the implemented model with a robot framework and demonstrate its utility for interactive object picking.

One limitation of the proposed system is potential discrepancies between our synthetic language templates and natural queries given arbitrarily by humans. In a relatively specialized domain (e.g., tabletop scenes of objects), this discrepancy can be alleviated by composing suitable modules based on language structure. 
However, the more one wants to cover generic content, the more refining the system architecture needs, which grants the work of modeling the domain and identifying useful linguistic priors a more cumbersome task, in which case end-to-end learning is favored. 
In the future, we would like to investigate the possibility of incremental acquisition of entity, attribute, and spatial concepts, e.g. by using a replay mechanism to update visual representations while tackling catastrophic forgetting. 
Additional directions ahead include extending the range of vision-and-language tasks tackled by the proposed framework, including VQA and visual reasoning.

\bibliography{main}
\bibliographystyle{collas2022_conference}
\newpage 

\appendix
\section{Visualization Examples of Model Inference}
\label{appendixA}
\begin{figure*}[h]
    \centering
    \vspace{3mm}
    \includegraphics[width=1\textwidth]{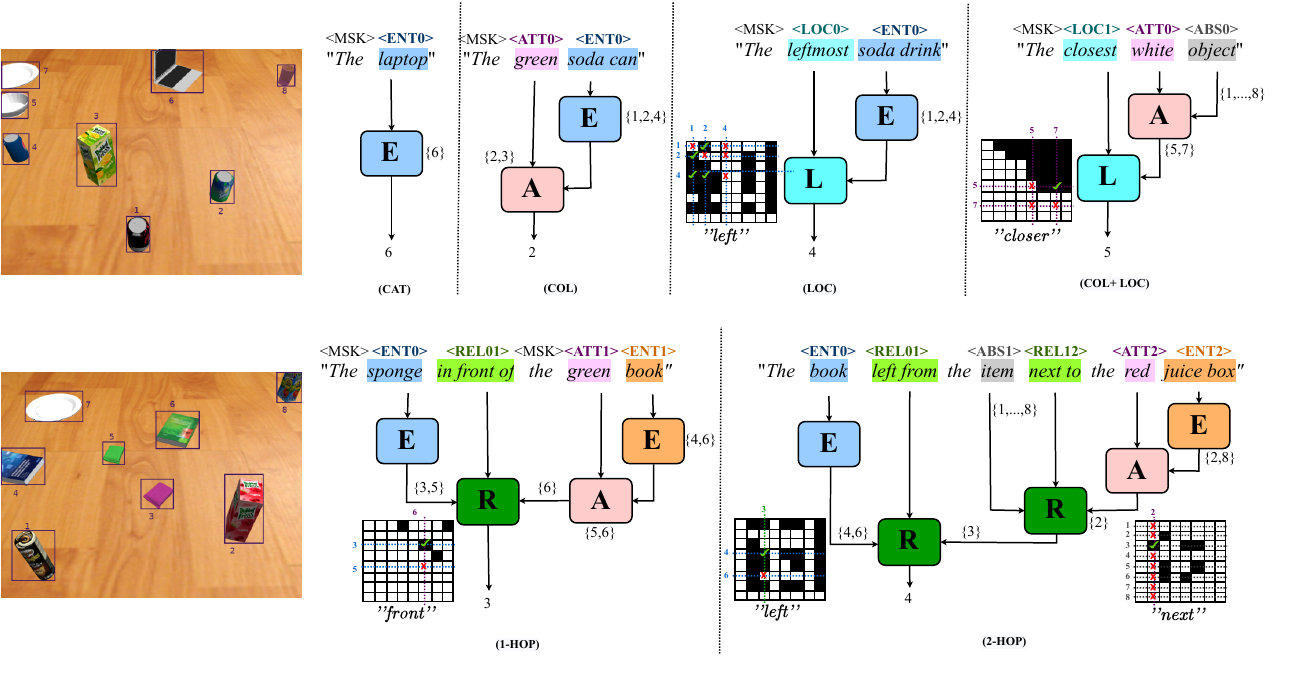}
    \caption{Output tags (illustrated on top of the input query) and corresponding module composition for inference. Examples include all different types of ambiguity resolution, from top-left to bottom-right: category (no ambiguity), color, location, color + location, mixed (1-hop) and mixed (2-hop)} .
    \label{fig:appA1}
\end{figure*}

\begin{figure*}[h]
    \centering
    \includegraphics[width=1\textwidth]{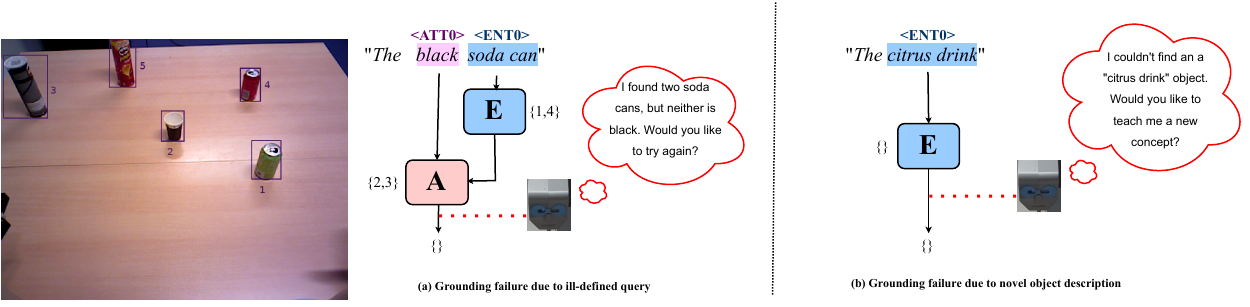}
    \caption{ Illustration of how adding exception traces  in the outputs of each module, combined with suitable responses, could extend the present framework to a dialogue setup, enabling users e.g. to repeat in case of ill-defined instructions (left) or even teach the robot a new concept (right).}
    \label{fig:appA2}
\end{figure*}

An important benefit of the implemented framework is that it enables full interpretability of the underlying reasoning steps taken to perform a grounding.
The compositional nature of the modules during test time allows us to inspect the output of each module and diagnose the source of error in case of failure.
In Fig.~\ref{fig:appA1} we demonstrate the predicted tags, the corresponding module composition and intermediate module outputs for grounding examples of all types of ambiguity resolution considered in this work. 
In Fig.~\ref{fig:appA2}, we illustrate how one could add exceptions in the outputs of the modules to capture specific failures and query them back to the human instructor. 
This allows extending the system to a dialogue setting, potentially enabling human users to incrementally teach the agent novel object-word pairs.
 
\section{Visual Grounding Evaluation in Simulation Experiment}
\label{AppendixB}
For entity recognition (category), we use a pre-trained \textit{MobileNet v2} network \citep{mobilenet} and fine-tune it on segmented crops from the synthetic scenes. 
We isolate the feature extraction layers and use them as the visual representation backbone for this round of experiments.
For attribute representation , since we only experiment with color, we encode each input crop with a $\mathbb{R}^{3 \times 256}$ color histogram vector, concatenated per channel. We have found experimentally that this representation is robust for both synthetic and real data.

In the feature-fusion baselines the input query is encoded into a single vector with a GRU encoder and concatenated with each entity (E) representation before passed to an MLP for ranking. 
We use pre-trained Glove \citep{glove} embeddings for word representation and a single bi-directional GRU layer with hidden size $256$. 
The MLP has two hidden layers of sizes $1024$, and $128$, respectively and rectified linear unit (ReLU) activation is used in both layers. We repeat our experiments but further concatenate each visual-query representation with attribute (A) and bounding box location (L) features. 

Additionally, we include modular approaches, based on CMN \citep{CMN}, where we keep the same visual representation but now use a soft language parser to decompose the input query into separate modules for subject-relation-object matching, potentially adding attribute and location modules explicitly, as in \citep{Mattnet}. We implement the parser as the same GRU network paired with three attention layers for each module and use hidden sizes of $256$ for the unitary and  $50$ for the pair-wise spatial relation matching networks. We train the model end-to-end in the weakly supervised setting, using the ground truth ROI index with a softmax loss over the final predictions per object.

For our hard parser-based model, we replace the soft parser with a neural module for word tagging and exploit the assumed triplet structure to compose modules in test time based on the predicted tags. 
First, we train the word tagger using a batch size of $64$ with a softmax loss over the vocabulary of tags we describe in Section~\ref{lang} (as e.g. in part-of-speech tagging).
As we train using synthetic data generated through templates, the ground truth tag annotations are taken from the template that generates the query. 
To train entity/attribute modules, we exploit the CLEVR-like functional program annotations of our generated dataset that provide us with execution traces for resolving each query. 
Such traces include annotations of which object indices and words are inputs/outputs to all modules separately. 
We extract and express these input/output indices as binary vectors (over the set of objects) and treat each object as a Bernoulli variable. 
We train the modules with a Binary Cross Entropy (per-logit) loss over the accumulated datasets for entity and attribute input/output pairs separately. 
For the spatial relation module $B$, since we have pair-wise scores for object pairs, we first flatten the pair-wise matrix to 1D and train with a BCE (per-logit) loss, using the flattened pair-wise matrix of the queried relation (again retrieved from the execution traces of our synthetic dataset) as supervision.
For the spatial module we use a batch size of $8$.
All models are implemented in PyTorch~\citep{pytorch} and trained using the AdamW optimizer~\citep{adamw}, with a batch size of $256$ unless stated otherwise, and an original learning rate of $5 \times 10^{-4}$ and a linear decay scheduler. 

We highlight that in the original CMN network the authors mentioned that their model is able to be trained using \textit{strong} supervision, meaning pair-wise annotations between the objects in the scene for the specific relation mentioned in the query. 
This ground truth signal is the same we use to train the spatial-language module $B$, as described above.
This is in fact a more relaxed notion of \textit{strong} supervision as meant in the context of this work, as in CMN all modules and the soft parser are trained jointly by this signal.
Instead, in our work we train each module independently.
When adding explicit attribute and location modules, the soft-parser baseline moves to the MattNet \citet{Mattnet} formalism, where all modules compute unitary scores and there is no pair-wise contextualization in any part of the network. 
That is because the CMN work only studies simple relation queries (without attributes/locations) and treats the subject / object as separate entities to be grounded. 
On the contrary, MattNet joins the relation and the object modules into a single relation module, removing the pair-wise computation between subject and object. As a result, the only variation of this baseline that can be trained in the strong supervision setting (in the CMN notion - meaning pair-wise annotations) is the $(E)+SoftParser$.
We confirm experimentally that using the same dataset size the ``strong'' version of this baseline achieves marginally same results as the weakly trained one, although converging must faster.


\section{Sim-To-Real Visual Domain Adaptation Experiment}
\label{AppendixC}
\begin{figure*}[h]
    \centering
    \includegraphics[width=1\textwidth]{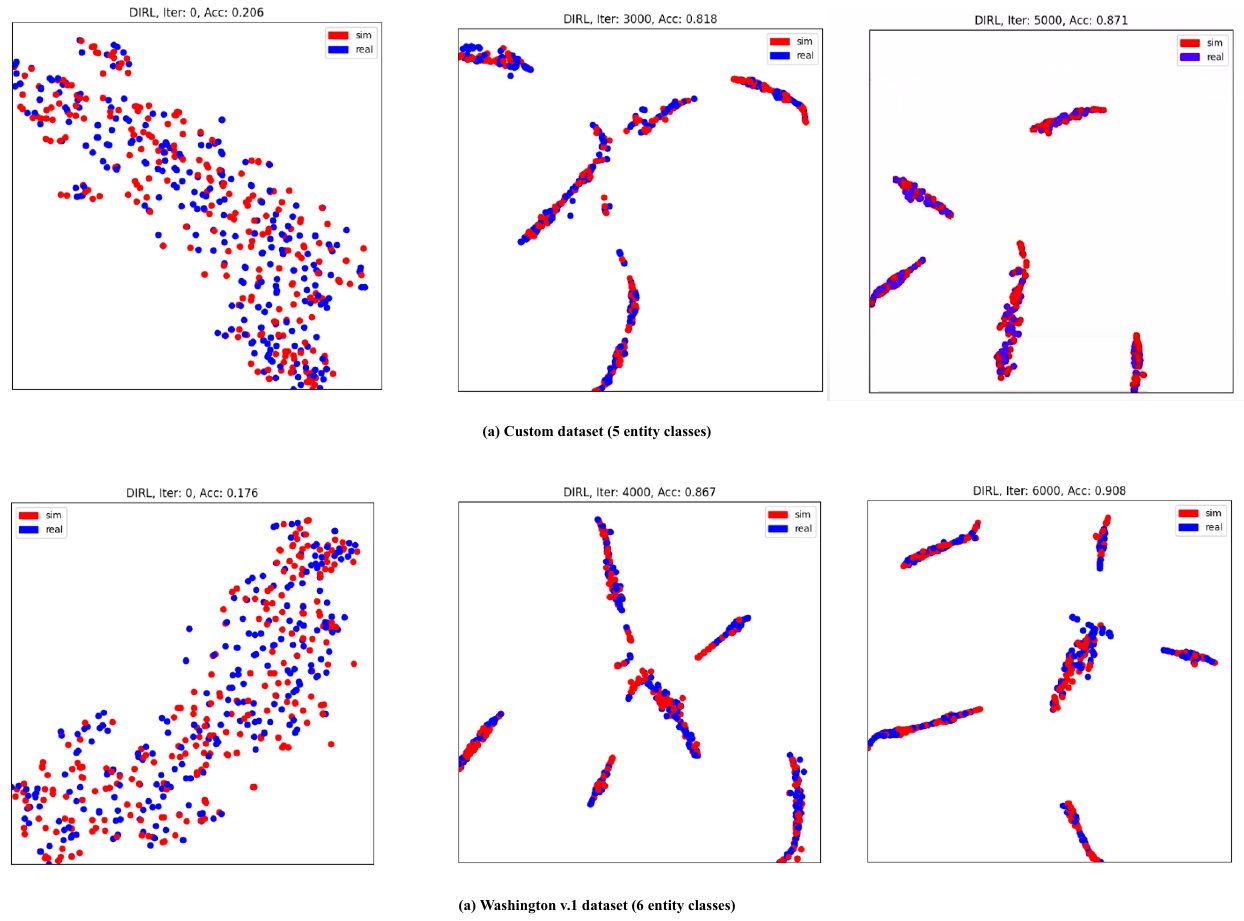}
    \caption{t-SNE projections of the embedding spaces learned by our CNN encoder during adversarial pre-training for Sim-To-Real adaptation. We visualize results for our Custom dataset (\textit{top}) and the Washington v.1 dataset (\textit{bottom}) during the start of training (\textit{"left"}), after 3-4 thousand iterations respectively (\textit{"middle"}) and after reaching maximum classification accuracy in the test split \textit{("right"}). We randomly sample $500$ images from each domain to generate the t-SNE projections.}
    \label{fig:tSNE}
\end{figure*}

For our experiments, we adapt the CNN encoder architecture of the digit experiments from the original DIRL paper, after we add an additional  convolutional-pooling-block in the input and fix image resolution to $75 \times 75$. 
We use $20$ labeled examples from the real domain and train using AdamW with a learning rate of $10^{-3}$. 
We split the batch size with $50\%$ sim and $50\%$ real data, while randomly shuffling the few labeled examples within the $25\%$ of the real part. 
We augment the labeled batch with random cropping and illumination distortions.
For the rest $75\%$, we sample unsupervised images with replacement.
We set the weights for all the constituent losses to $1$ and use a margin of $0.6$ for the triplet KL loss. 
We skip the pseudo-labeling process during training, as it experimentally did not provide improvements.
Fig.~\ref{fig:tSNE} presents the incremental progression of the learned representations towards domain-invariant representations for both real datasets.

\begin{figure*}[t]
    \centering
    \includegraphics[width=1\textwidth]{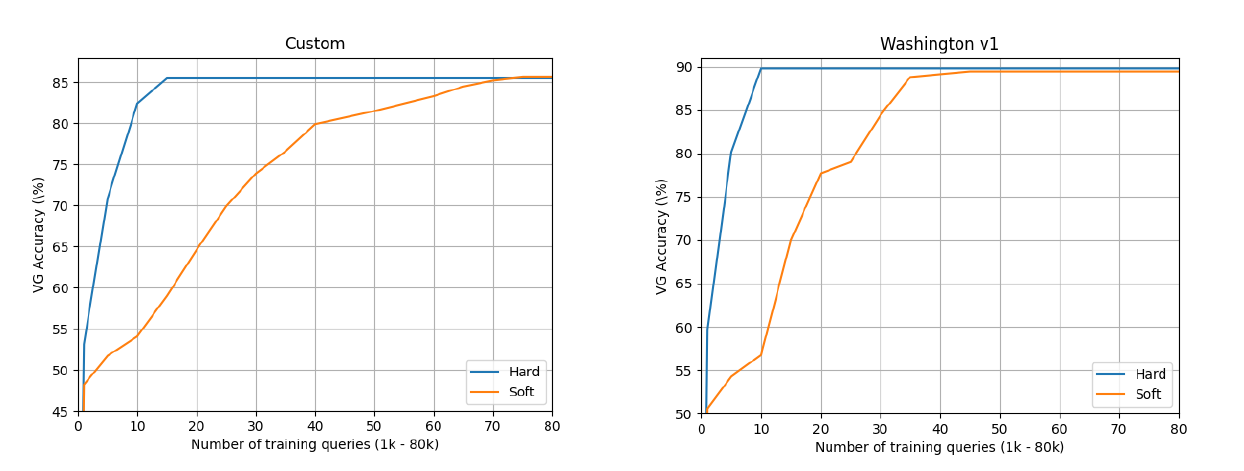}
    \vspace{-5mm}
    \caption{Top-1 predicted accuracy vs. number of synthetic image-query samples used for fine-tuning both models after adapting the visual representations. Results are presented for our Custom (\textit{left}) and the Washington-v.1 (\textit{right}) dataset.
    The test splits include only "simple" cases of queries (category, location, attribute and entity-only relation).
    Even though both models eventually reach the same performance, our hard-parser-based model is much more data efficient, due to decoupling training for language parsing and the matching modules}
    \label{fig:accVSsize}
    \vspace{-2mm}
\end{figure*}

Additionally, we perform a comparative study of visual grounding performance of our model versus the soft-parser-based baseline, for different dataset sizes. 
As we have already highlighted in Section~\ref{sim2real}, the soft-parser baseline requires fine-tuning of all modules in an end-to-end fashion, including the language parser. 
In absence of real natural language annotations for the scenes in the target domains, one could use the synthetic queries and train the model from scratch after only adapting the visual representations.
We perform this experiment for different volumes of image-query samples and compare with our hard-parser-based model. 
We note that as in the original experiment, our model here is not trained end-to-end, but the parser and the matching modules are trained independently and combined only for evaluation in the real datasets. 
In order to provide fair comparisons, we exclude from the test splits of the target datasets all queries that the soft-parser baseline cannot handle (mixed 1-hop, 2-hop).
Fig.~\ref{fig:accVSsize} illustrates the top-1 accuracy of both datasets for different dataset sizes.

We observe that given enough data, the soft-parser is indeed capable or reaching parallel performance to our model (in the excluded split). 
However, our model is a lot more data-efficient, as we observe from our results.
Notably, we validate that using only $20k$ image-query pairs ($25\%$ of the training data) our word tagger network achieves $99.7\%$ token-level and $95.1\%$ sentence-level accuracy when evaluated using the ground truth tags of the synthetic dataset. 
This is a benefit from utilizing explicit supervision for training the language module, versus training it end-to-end with the rest of the network.
Even though additional annotations are generally considered a bottleneck, we highlight that since we train with language data \textit{only} in simulation, this limitation is alleviated by the fact that in with the simulator we can extract such rich annotations "for-free", in large scale and in a controlled fashion.

\section{Comparisons with Zero-Shot Large-Scale Transformers}
 \label{appendixD}
A well established methodology in modern machine learning focuses around using large-scale representation learners, usually transformers, most popularly pre-trained in a self-supervised manner in massive datasets (i.e. \textit{Foundation models}).
The model is then fine-tuned for a downstream task or applied in a zero-shot setting.
This methodology is also popular in vision-and-language learning, with an array of different transformer architectures and pre-training objectives (\citet{vilbert, visual-bert, UNITER, 12in1}).
Compared to modular approaches such as ours, that are very closely tight to the application domain, such pre-trained methods have shown to provide tremendous generalization capabilities in arbitrary downstream tasks.
This benefit is due to the representational strength of the pre-trained embeddings, as the model is exposed to massive amounts of data.
In order to explore trade-offs between such methods and the one we propose in this work, we evaluate two zero-shot transformer approaches in our synthetic dataset after adapting them for the visual grounding task.

\begin{wraptable}{r}{0.3\textwidth}
    \vspace{-5mm}
    \centering {
    \resizebox{\linewidth}{!}{
    \begin{tabular}{lcc}
     \textbf{Model} & \textbf{Visual} & \textbf{Spatial} \\
     \toprule
     \emph{CLIP-RN50x4} & $90.3$ & $50.8$ \\
     \emph{CLIP-RN50x16} & $95.1$ & $52.8$ \\
     \emph{CLIP-RN50x64} & $92.6$ & $49.8$ \\
     \emph{CLIP-ViT-B/32} & $88.0$ & $47.5$ \\
     \emph{CLIP-ViT-B/16} & $91.4$ & $48.7$ \\
     \toprule
     \emph{ALBEF-ITM} & $89.8$ & $48.4$  \\
     \emph{ALBEF-ITC} & $73.3$ & $48.4$ \\
     \toprule
     \emph{ours} & $95.6$ & $96.9$ \\
    \end{tabular}}}
    \caption{Accuracy (\%) on our synthetic dataset for the image-text matching task. }
    \label{tab:clip}
    \vspace{-3mm}
\end{wraptable}
We experiment with two models, CLIP \citep{CLIP} and ALBEF \citep{ALBEF}.
CLIP employs one image-only and one text-only encoder and uses a contrastive loss on top of them to learn representations that are closer in the embedding space for matched image-text pairs. Similarly, ALBEF further employs a multi-modal encoder and pre-trains with three losses, a) an image-text contrastive loss (ITC) such as CLIPs, b) an image-text matching loss (ITM) in the output of the multi-modal encoder, and c) a masked language modeling loss for the text encoder. 

In order to evaluate the models for zero-shot visual grounding, we replicate the \textit{text-pair} experiment introduced in the ReCLIP \citep{reclip} paper for the CLEVR dataset. In this setup, a false version of the input query is generated, both for spatial (e.g. positive: \textit{``bowl behind flashlight}, negative: \textit{``bowl in front of flashlight''}) and for visual only queries (e.g. positive \textit{``a blue bowl and black flashlight''}, negative: \textit{``a red bowl and a black flashlight''}). 
Both queries are passed to the model and we measure accuracy as the percentage of queries that the positive version scores higher than the negative. 
For CLIP, we experiment with different ResNet-50 \citep{resnet} and Visual Transformer (ViT) \citep{ViT} 
encoders. Following the original paper, we experiment with ALBEF using both the ITC and the ITM scores for computing similarity scores for image-text pairs.
We present results separated in two splits, namely: a) \textbf{Visual}, where the query contains only object entity and color and b) \textbf{Spatial}, where the query is a 1-hop entity-only relation. Results are summarized in Table~\ref{tab:clip}. It can be seen that both CLIP and ALBEF perform well in the visual version of the task but poorly in relation queries. The evaluation split contains $895$ scene instances.

Our experiment reproduces the findings of the ReCLIP paper, namely that even though pre-trained models perform well on the visual task, they cannot comprehend spatial relations. The authors alleviate this by extending the model to also include a ROI proposal generator and a rule-based relation extractor.
With this formulation the grounding model becomes very similar to ours, with the exception of using a hand-crafted relation module and CLIP/ALBEF for computing visual matching scores, instead of a module trained in a supervised fashion in the domain.
With the insight of this result, combined with additional benefits of our approach, namely: a) computational efficiency, as our  method has less than $0.2\%$ the parameter count of CLIP and b) interpretability, due to full decoupling of modules in our approach, grants our model favourable for application. 
We highlight however that this is expected, as our model is designed with domain knowledge about the task (grammatical structure of relations in natural language) injected in its architecture. 
Foundation models still offer a strong baseline, especially considering their zero-shot capabilities, although still lacking when compared with domain-specific architectures for a given task.

\end{document}